\documentclass[lettersize,journal]{IEEEtran}
\usepackage{amsmath,amsfonts,bm}
\usepackage{amsmath}
\usepackage[linesnumbered,ruled,commentsnumbered]{algorithm2e}
\usepackage{array}
\usepackage{booktabs}
\usepackage[caption=false,font=normalsize,labelfont=sf,textfont=sf]{subfig}
\usepackage{tabularx}
\usepackage{siunitx}

\usepackage{diagbox}
\usepackage{makecell}
\usepackage{textcomp}
\usepackage{stfloats}
\usepackage{url}
\usepackage{verbatim}
\usepackage{graphicx}
\usepackage{color}
\usepackage{nomencl}

\usepackage{svg}

\usepackage{hyperref}
\newcolumntype{C}{>{\centering\arraybackslash}X}
\hypersetup{
colorlinks=true,
linkcolor=blue,
filecolor=purple,      
urlcolor=blue,
citecolor=green,
}
\usepackage{cite}
\usepackage{url}

\makenomenclature
\begin{document}

\title{Escaping Stability-Plasticity Dilemma in Online Continual Learning for Motion Forecasting via Synergetic Memory Rehearsal}

\author{Yunlong Lin,  Chao Lu,~\IEEEmembership{Member,~IEEE}, Tongshuai Wu, Xiaocong Zhao, Guodong Du, Yanwei Sun, Zirui Li,~\IEEEmembership{Member,~IEEE},  Jianwei Gong,~\IEEEmembership{Member,~IEEE}
\thanks{This research is supported by National Science and Technology Major Project nder Grant 2022ZD0115503 and the National Natural Science Foundation of China under Grant 52372405. (Corresponding authors: Zirui Li and Jianwei Gong)}
\thanks{Yunlong Lin,  Chao Lu, Tongshuai Wu, Guodong Du, Zirui Li and Jianwei Gong, are with the School of Mechanical Engineering, Beijing Institute of Technology, Beijing 100081, China. (Email: jacklyl.bit@gmail.com; chaolu@bit.edu.cn; wutongshuai84@gmail.com; guodongdu\_robbie@163.com; ziruili.work.bit@gmail.com; gongjianwei@bit.edu.cn)}
\thanks{Chao Lu and Jianwei Gong are also with Zhengzhou Research Institute, Beijing Institute of Technology, Zhengzhou 450000, China.}
\thanks{Zirui Li is also with the School of Mechanical and Aerospace Engineering, Nanyang Technological University, Singapore, 639798.}
\thanks{Xiaocong Zhao is with Key Laboratory of Road and Traffic Engineering, Ministry of Education, Tongji University, Shanghai, China (Email: zhaoxc@tongji.edu.cn).}
\thanks{Yanwei Sun is with Zhengzhou Nissan Automobile Co., Ltd., Zhengzhou 450046, China (Email: sunyanwei@zznissan.com.cn)}
}


\maketitle

\begin{abstract}
Deep neural networks (DNN) have achieved remarkable success in motion forecasting. However, most DNN-based methods suffer from catastrophic forgetting and fail to maintain their performance in previously learned scenarios after adapting to new data. Recent continual learning (CL) studies aim to mitigate this phenomenon by enhancing \emph{memory stability} of DNN, i.e., the ability to retain learned knowledge. Yet, excessive emphasis on the memory stability often impairs \emph{learning plasticity}, i.e., the capacity of DNN to acquire new information effectively. To address such stability-plasticity dilemma, this study proposes a novel CL method, synergetic memory rehearsal (SyReM), for DNN-based motion forecasting. SyReM maintains a compact memory buffer to represent learned knowledge. To ensure memory stability, it employs an inequality constraint that limits increments in the average loss over the memory buffer. Synergistically, a selective memory rehearsal mechanism is designed to enhance learning plasticity by selecting samples from the memory buffer that are most similar to recently observed data. This selection is based on an online-measured cosine similarity of loss gradients, ensuring targeted memory rehearsal. Since replayed samples originate from learned scenarios, this memory rehearsal mechanism avoids compromising memory stability. We validate SyReM under an online CL paradigm where training samples from diverse scenarios arrive as a one-pass stream. Experiments on 11 naturalistic driving datasets from INTERACTION demonstrate that, compared to non-CL and CL baselines, SyReM significantly mitigates catastrophic forgetting in past scenarios while improving forecasting accuracy in new ones. The implementation is publicly available at \href{https://github.com/BIT-Jack/SyReM}{https://github.com/BIT-Jack/SyReM}. 

\end{abstract}


\section{Introduction}

\begin{figure}[tp]
      \centering
      \includegraphics[scale=1.0]{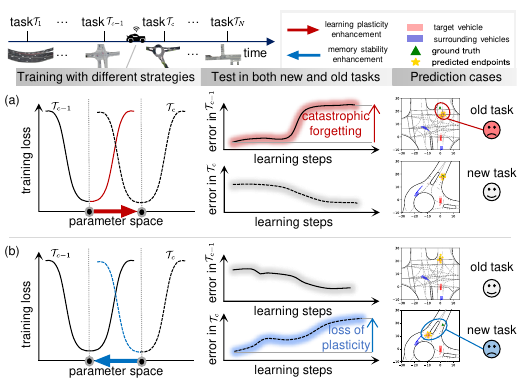}
      \captionsetup{font={small}}
      \caption{Stability-plasticity dilemma in continual learning tasks of motion forecasting. (a) Optimizing for the new task may overwrite parameters that are important to old tasks, leading to catastrophic forgetting in old tasks. (b) Excessive memory stability may restrict the adaptation of DNN to new tasks, causing a loss of learning plasticity.}
      \label{fig_intro}
\end{figure} 

\IEEEPARstart{D}{eep} neural networks (DNN) have been successfully applied in motion forecasting~\cite{2025-GCN0traj-nnls}, which is important to the safe planning in autonomous driving (AD)~\cite{2025Haochen-TPAMI}. As naturalistic scenarios evolve dynamically, AD systems need to adapt and remember incrementally encountered scenarios throughout their lifespans~\cite{kudithipudi2022biological-nature}. To enhance generalization performance across different scenarios, most motion forecasting methods train DNN using a vast amount of data, trying to cover as many scenarios as possible~\cite{korbmacher2022deep-review}. However, real-world scenarios are infinite. The performance of DNN-based motion forecasting cannot be guaranteed in scenarios that have not been explicitly encountered during training. Meanwhile, simultaneously maintaining massive data from potentially unlimited scenarios overwhelms the finite storage and computation resources of AD systems~\cite{mozaffari2020deep}. In such cases, continuously updating the DNN as new data arrives is a potentially feasible paradigm for DNN-based motion forecasting methods to obtain consistently accurate predictions across different scenarios~\cite{li2023continual-survey}. Nevertheless, within this paradigm, DNN-based methods may suffer severe performance degradation in previously learned scenarios after adapting to new ones, a phenomenon known as catastrophic forgetting~\cite{mccloskey1989catastrophic}. This phenomenon arises mainly because when learning new scenarios with shifted data distributions, the trainable parameters of the DNN tend to be optimized for the new data distribution. Such distributional shift-guided parameter updates overwrite the knowledge essential for previously learned scenarios~\cite{2025pami-forgetting}.

As a widely used solution, continual learning (CL) is proposed to address the catastrophic forgetting of DNN~\cite{wang2023incorporating-nmi}. Specifically, CL aims to enable DNN to learn sequential tasks with no or limited access to old data, while achieving strong performance on the test sets of all encountered tasks~\cite{lesort2020cl-robot-survey}. This goal of CL essentially boils down to resolving the stability-plasticity dilemma~\cite{lu2025rethinking}: \emph{memory stability} refers to the capability of avoiding catastrophic forgetting on past tasks, and \emph{learning plasticity} denotes the capacity to adapt to new tasks efficiently~\cite{lyle2023understanding}, yet the two compete for the same parameter space in DNN~\cite{TPAMI2024-comprehensive-cl}. As illustrated in Fig.~\ref{fig_intro}(a), the DNN updates its parameters to fit data distributions from the new task $\mathcal{T}_c$, which risks eroding the parameter configurations that underpin memory stability. Consequently, catastrophic forgetting happens, leading to poor performance in old tasks (e.g., task $\mathcal{T}_{c-1}$). Conversely, rigidly preserving parameters to maintain memory stability would stifle the necessary adjustments for learning plasticity, impeding the model from effectively adapting to new tasks (Fig.~\ref{fig_intro}(b)). This inherent competition gives rise to the stability-plasticity dilemma.

From the aspect of DNN-based motion forecasting, the CL tasks are defined as predicting trajectories or behaviors of road users (e.g., vehicles, pedestrians, and cyclists) across different scenarios~\cite{ma2021continual, lin2024continual}. Recent studies have investigated CL tasks of motion forecasting~\cite{ma2021continual, wu2022continual-ped, lin2024continual, 2022-EWC-ped-traj, bao2023lifelong, lin2025h2c}, where CL tasks with corresponding datasets incrementally arrive as a sequence. Usually, various datasets collected from different locations are used to construct the CL tasks for motion forecasting. Each time a task arrives, the corresponding dataset of this task (termed the current task) is fully available. Under the assumption of limited data usage from past tasks, these studies attempted to alleviate catastrophic forgetting and aimed to minimize the averaged prediction errors in test sets of all encountered tasks. Metrics to measure the memory stability and overall performance were highlighted in their evaluations~\cite{ma2021continual, wu2022continual-ped, lin2024continual}. Nevertheless, the methodology and the validation for DNN's learning plasticity are almost ignored. Such ignorance could be due to their offline CL manner, where the DNN can be trained for any epoch based on the fully available dataset from the current task. Thus, attaining accurate predictions on new tasks is hardly a hurdle, and these works instead prioritize overcoming catastrophic forgetting merely. However, allocating exclusive time to offline training, a process that imposes heavy demands on large datasets and entails lengthy training cycles, is scarcely viable in AD deployments with stringent real-time requirements~\cite{2024-CL-object_recog}.

By contrast, online CL trains DNN using a one-pass data stream~\cite{TPAMI2024-comprehensive-cl}, which could be a more practical paradigm for motion forecasting in AD~\cite{online2024gnn-nnls}. A recent study has empirically shown the poor memory stability of state-of-the-art DNN-based motion forecasting models when learning a one-pass data stream in the online CL paradigm~\cite{lin2023rethinking}. On the other hand, recent theoretical findings have demonstrated that DNN-based methods in CL settings gradually lose learning plasticity until no better than a shallow network~\cite{dohare2024plasticity-nature}. Directly applying established CL-based motion forecasting methods could further harm the learning plasticity in such an online CL paradigm~\cite{lyle2023understanding}. As a result, most DNN-based motion forecasting methods still get trapped in the dilemma of trading off memory stability and learning plasticity. Escaping the stability-plasticity dilemma of online CL for motion forecasting is significant but challenging for the development of high-level AD systems. To our best knowledge, few studies have investigated this research field.

To fill this research gap, this study proposes a novel CL method, termed Synergetic Memory Rehearsal (SyReM), for DNN-based motion forecasting to escape the stability-plasticity dilemma. First, we design a memory buffer that encapsulates knowledge learned from samples of previously encountered tasks, serving as a foundational component in the SyReM. For memory stability, the primary goal is to prevent the erosion of knowledge from prior tasks when the model adapts to new data distributions. Given that DNN parameter updates rely on gradient descent, SyReM employs a gradient projection strategy based on the memory buffer. This strategy dynamically adjusts the gradient descent process within the online CL paradigm, constraining the incremental increase of the averaged loss over the memory buffer. Since the learned knowledge is represented by the buffer, this constraint mitigates the forgetting of historical tasks as the model adapts to new data distributions. Synergistically, an online similarity measurement-based rehearsal strategy is developed for learning plasticity. As the objective is to enhance the adaptive efficiency for new tasks by leveraging relevant prior knowledge, the strategy selects samples from the memory buffer that are most similar to the latest observed samples in the new task and rehearses them to the DNN at each training step. This process facilitates the transfer of useful prior knowledge to boost adaptation to new tasks. Notably, since the rehearsed samples are directly drawn from the memory buffer, they inherently avoid conflicting with the previously learned knowledge, thereby supporting memory stability while enhancing plasticity. The main contributions of this paper are summarized as follows:
\begin{itemize} 
    \item[1)] A novel CL method, SyReM, is proposed to resolve the stability-plasticity dilemma in DNN-based motion forecasting. It guarantees memory stability by constraining increments in the average loss over learned tasks, while synergistically enhancing learning plasticity through a selective memory rehearsal mechanism that leverages prior knowledge.
    \item[2)] Two core mechanisms of SyReM are tailored for the online CL paradigm, where sequential task data arrives as a one-pass stream without multi-epoch training on static task batches. First, an online memory buffer is constructed via reservoir sampling to capture the distribution of learned samples. Second, a selective rehearsal strategy compares cosine similarities of loss gradients between old and new samples, enabling real-time guidance for memory rehearsal at each learning step.
    \item[3)] Comprehensive experiments are conducted using diverse metrics to systematically evaluate memory stability and learning plasticity. Experimental results demonstrate that SyReM outperforms both CL and non-CL baselines in simultaneously enhancing stability and plasticity. Beyond addressing the dilemma, SyReM exhibits superior zero-shot generalization. Ablation studies further validate the core mechanism by analyzing similarity measurements of rehearsed samples, confirming the effectiveness of the proposed strategy.
\end{itemize}

The remainder of this article is organized as follows. Section \ref{Section-II} first introduces the related works to distinguish this study from previous works. Then, online CL tasks for motion forecasting are formulated in Section \ref{Section-III}. The proposed method is detailed in Section \ref{Section-IV}. Next, experimental settings, results, and discussion are presented in Section \ref{Section-V}, respectively. Finally, the conclusion and future works are summarized in Section \ref{Section-VI}.

\section{Related Works}\label{Section-II}

This work seeks to enable DNN-based motion forecasting models to address the stability-plasticity dilemma in CL. In this section, existing studies on DNN-based motion forecasting are first reviewed to provide the necessary background. Next, the fundamental concepts and mainstream strategies of CL are introduced. Besides, to contextualize the dilemma and further delineate our contributions, related paradigms, including transfer learning (TL) and meta-learning, are briefly surveyed.

\subsection{Deep Neural Networks-based Motion Forecasting}
DNN-based motion forecasting predicts future trajectories of road users (e.g., vehicles, pedestrians, and cyclists) from their historical motion states and scene context~\cite{mozaffari2020deep, korbmacher2022deep-review}. Early studies used recurrent neural networks such as long short-term memory and gated recurrent unit to learn temporal patterns~\cite{2021-LSTM-traj-nnls}. To improve multimodality, variational autoencoders and generative adversarial networks frameworks are introduced~\cite{gupta2018socialGAN}. Such probabilistic models generate diverse plausible trajectories with uncertainty estimation. To capture spatial features, convolutional neural networks processed high-definition maps or bird’s-eye-view images~\cite{conv2025-nnls}. Recent studies have leveraged graph neural networks~\cite{online2024gnn-nnls, 2024-goal-traj-nnls} and Transformer architectures~\cite{2024-graphormer-ped-traj-nnls} or attention mechanisms~\cite{li2024UQnet} to enhance interaction reasoning among agents. However, most established DNN-based methods were evaluated using test data that is independent and identically distributed (i.i.d.) with the training data. The struggle with the adaptation to out-of-distribution limits their applicability in evolving scenarios.

\subsection{Continual Learning}

CL is also termed as lifelong learning~\cite{kudithipudi2022biological-nature} or incremental learning~\cite{2018incremental-nnls}. CL allows sequential task learning without forgetting past knowledge~\cite{TPAMI2024-comprehensive-cl}. Common strategies include regularization-based~\cite{2018-R-EWC}, architecture-based~\cite{feng2024-arc-tnnls}, and replay-based~\cite{2022-triplemem-CL-nnls}. Replay-based strategies are widely used in online CL by adapting to streaming inputs with bounded memory~\cite{2025-onlineCL-nnls}. 
Replay-based strategies store past examples for memory rehearsal~\cite{lesort2020cl-robot-survey}. Generative replay synthesizes pseudo-samples to approximate learned knowledge from past tasks~\cite{li2021-incre-gr}. During training, the generated pseudo-samples are interleaved with new samples to maintain memory stability. Nevertheless, most replay-based methods favor the memory stability on past tasks at the expense of learning plasticity for new tasks~\cite{dohare2024plasticity-nature}.

There are a few paradigms aiming to enhance the learning plasticity for DNN, including TL~\cite{tl2023-nnls}, and meta-learning~\cite{2023-meta}. TL usually pre-trains models on large source datasets and fine-tunes them on target tasks. Although the learning plasticity can be enhanced by TL with limited data, most TL-based models suffer from domain mismatch and catastrophic forgetting. Meanwhile, TL requires access to target data during training and may not continuously handle streaming scenarios~\cite{lu2019transfer-TITS, emami2022rc-transfer}. Alternatively, meta-learning learns task-agnostic initializations for fast adaptation~\cite{2022-meta-learing-survey}. However, the meta-learning relies on diverse meta-training tasks and exhibits high computational overhead. In summary, although TL and meta-learning have demonstrated progress in improving the learning plasticity of DNN, these paradigms might fail to reconcile the stability with plasticity and lack mechanisms for continuous updating. Trading off memory stability and learning plasticity is a critical trend for improving the practicability of DNN-based motion forecasting.

\section{Problem Formulation}\label{Section-III}

\begin{figure*}[tp]
      \centering
      \includegraphics[scale=1.0]{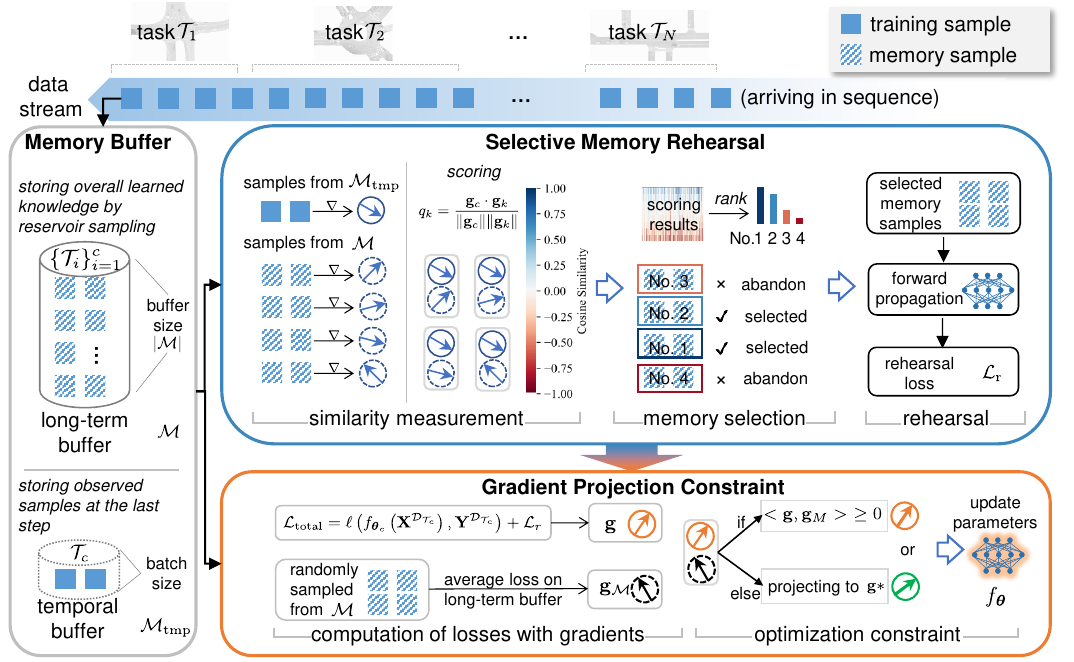}
      \captionsetup{font={small}}
      \caption{The proposed SyReM method. The memory buffer contains a long-term buffer $\mathcal{M}$ and a temporal buffer $\mathcal{M}_\text{tmp}$. First, overall knowledge from the online data stream is maintained in $\mathcal{M}$ via reservoir sampling, and the observed samples at the last learning step are temporarily stored in $\mathcal{M}_\text{tmp}$. Then, similarity scores are computed by comparing the gradient cosine similarity between $\mathcal{M}_\text{tmp}$ and candidates from $\mathcal{M}$. Memory samples with the higher scores are selected for the rehearsal, which enhances the learning plasticity. Finally, the gradient projection employs an inequality constraint for the DNN update to further improve memory stability.}
      \label{fig_method}
\end{figure*}

Motion forecasting for AD aims to estimate the future positions or behavior intentions of road users such as vehicles, pedestrians, and cyclists. In this context, target agents are defined as road users under prediction. Surrounding agents (SAs) are defined as other road users that can influence target agents (TAs) through physical interactions or environmental constraints. DNN-based motion forecasting models leverage the historical motion states of both TAs and SAs to achieve spatiotemporally accurate predictions.

Formally, let $t_{\text{c}}$ denote the current time step within the observation horizon $t_{\text{obs}}$. The input to a DNN-based motion forecasting model is represented as $\mathbf{X}=[\mathbf{X}_{t_{\text{c}}-t_{\text{obs}}+1}, ..., \mathbf{X}_{t_{\text{c}}}]$, where $\textbf{X}_{t}$ contains motion states (e.g., positions and velocities) of the TAs and SAs, and map features (e.g., road boundary and centerline of lanes) at time $t$. Let $f_{\boldsymbol{\theta}}$ denote the DNN-based model paramterized by $\boldsymbol{\theta}$. Given a forecasting time window $t_\text{pred}$, the output $f_{\boldsymbol{\theta}}(\mathbf{X})$ estimates the future motion states $\mathbf{Y}$ of the target agent at time $t_\text{c}+t_\text{pred}$.

In the online CL setting, the model $f_{\boldsymbol{\theta}}$ will encounter $N\in\mathbb{Z}^{+}$ tasks $\{\mathcal{T}_{1},...,\mathcal{T}_{N}\}$ sequentially. Let $i$ denote the index of tasks, and let $j$ denote the index of samples. $\forall{i} \in [1, N]$, each task $\mathcal{T}_{i}$ contains $N^{\text{tr}}_{\mathcal{T}_{i}}$ training samples drawn from an i.i.d. distribution $\mathcal{D}_{\mathcal{T}_{i}}$. Toward the online characteristic~\cite{TPAMI2024-comprehensive-cl}, the corresponding training samples of all CL tasks arrive as a one-pass data stream:
\begin{equation}
    \mathcal{S}=\{{(\mathbf{X}, \mathbf{Y})}_{j}\sim\mathcal{D}_{\mathcal{T}_i} | 1 \le i \le N, 1 \le j \le N^{\text{tr}}_{\mathcal{T}_i} \}\label{eq_data_stream}
\end{equation}

We use $\mathcal{T}_{c}$ to denote the current task that the model is learning, where the subscript $c \in [1, N]$. Then, the past tasks are defined as the set of previously learned tasks $\{\mathcal{T}_{i}\}_{i=1}^{c-1}$. Since memory stability refers to the ability to retain learned knowledge in past tasks, the training loss of memory stability can be formulated as: 
\begin{equation}
    \mathcal{L}_{\text{s}} = \sum_{i=1}^{c-1}\mathbb{E}_{(\mathbf{X},\mathbf{Y})\sim\mathcal{D}_{\mathcal{T}_{i}}}\left( \ell\left(f_{\boldsymbol{\theta}_{i}}\left( \mathbf{X}\right), \mathbf{Y}\right)\right)
    \label{eq_stability}
\end{equation}
where $\ell(\cdot)$ is the loss function, and $\mathbb{E}$ denotes the mathematical expectation. The $f_{\boldsymbol{\theta}_{i}}$ represents the motion forecasting model trained on task $\mathcal{T}_i$. By contrast, learning plasticity refers to the ability of DNN to learn new tasks. This necessitates utilizing learned knowledge from past tasks to obtain accurate predictions in the current task efficiently. Thus, the loss of learning plasticity is represented as:

\begin{equation}
   \mathcal{L}_{\text{p}} = \mathbb{E}_{(\mathbf{X},\mathbf{Y})\sim\mathcal{D}_{\mathcal{T}_{c}}}\left( \ell\left(f_{\boldsymbol{\theta}_{c}}\left( \mathbf{X}\right), \mathbf{Y}\right)\right)
    \label{eq_plasticity}
\end{equation}

However, the excessive stability may hinder efficient adaptation to the new task, while over-focusing on the plasticity may aggravate catastrophic forgetting in past tasks. 
The stability-plasticity dilemma arises from the difficulty of achieving the objectives \eqref{eq_stability} and \eqref{eq_plasticity} simultaneously:

\begin{equation}
        \boldsymbol{\theta}_{1:c}^{*} = \arg\min_{\boldsymbol{\theta}_{1:c}} \mathcal{L}_{\text{s}}+\mathcal{L}_{\text{p}}
    \label{eq_dilemma}
\end{equation}
Here, we use $\boldsymbol{\theta}_{1:c}$ to denote parameters of the model that is trained from the first task $\mathcal{T}_1$ to the current task $\mathcal{T}_c$. The aim is to find the optimal parameters $\boldsymbol{\theta}^{*}_{1:c}$, minimizing both the loss of memory stability and learning plasticity.

\section{Synergetic Memory Rehearsal for Online Continual Learning}\label{Section-IV}
This study proposes a method, SyReM, for DNN-based motion forecasting to escape the stability-plasticity dilemma. The overall schematic of the SyReM is demonstrated in Fig.~\ref{fig_method}. First, a memory buffer is designed to partially store learned samples from the online data stream. Then, as a core component of SyReM, the selective memory rehearsal aims at enhancing learning plasticity by leveraging samples from the memory buffer on a selective basis through a similarity measurement. Combined with the rehearsal loss output by the selective memory rehearsal module, the gradient descent on the total loss is constrained for the guarantee of memory stability via a gradient projection strategy. More details are presented as follows.

\subsection{Memory Buffer for Online Data Stream}
The proposed methodology is built on a replay-based CL mechanism, representing learned knowledge by a subset of observed samples. For the representation of knowledge from past tasks, we use a long-term memory buffer $\mathcal{M}$ to dynamically maintain a subset of observed samples from the data stream, called memory samples. Due to the resource efficiency of CL, the amount of memory samples in the long-term buffer, i.e., the buffer size denoted as $|\mathcal{M}|$, is required to be far less than the total number of training samples~\cite{lesort2020cl-robot-survey}:
\begin{equation}
    |\mathcal{M}| \ll \sum_{i=1}^{N}N_{\mathcal{T}_i}^{\text{tr}}\label{eq_data_limit}
\end{equation}

With the limited memory resources, the long-term buffer is designed to approximate the overall distribution of all learned samples. Equiprobable sampling is an effective strategy for obtaining an unbiased estimation of the entire training set. However, in online CL, samples become sequentially available through the data stream defined in \eqref{eq_data_stream}, where the total length of the data stream, $|\mathcal{S}|$, remains unknown. This lack of prior knowledge about $|\mathcal{S}|$ renders commonly used algorithms such as simple random sampling or systematic sampling inapplicable to such an online learning paradigm. To address this challenge, we dynamically update the long-term buffer using reservoir sampling~\cite{vitter1985random-rsvr}. First, the memory buffer is initialized to an empty set. Before full, every observed sample is sequentially stored in the buffer. After that, the newly encountered sample may replace one of the stored samples in the buffer. 
As depicted in Algorithm~\ref{al_1}, $\forall k > |\mathcal{M}|$, the $j\textsuperscript{th}$ observed sample is selected as the candidate with the probability of $|\mathcal{M}|/k$. Once the $j\textsuperscript{th}$ sample becomes the candidate, one of the previously stored samples will be replaced by the candidate with the probability of $1/|\mathcal{M}|$. Let $\mathcal{M}(i)$ denote the $i$\textsuperscript{th} sample stored in the memory buffer. The update step is implemented via a conditional replacement:
\begin{equation}
    \mathcal{M}(r) = \begin{cases} 
(\mathbf{X}, \mathbf{Y})_k, & \text{if } r \le |\mathcal{M}|,  r \sim \mathrm{Uniform}(1,k) \\
\mathcal{M}(r), & \text{otherwise.}
\end{cases}
\end{equation}
where $r\in\mathbb{Z}^{+}$ is the randomly generated index. As a result, the probability of being stored in the buffer is $|\mathcal{M}|/|\mathcal{S}|$ for each sample in $\mathcal{S}$.

\begin{algorithm}[tp]
\SetAlgoNlRelativeSize{0}  
\SetNlSty{textbf}{}{} 
\SetAlgoNlRelativeSize{-1} 
\SetCommentSty{textnormal} 
\SetKwComment{tcp}{$\triangleright$}{}

\caption{Sampling Strategy for Memory Buffer}\label{al_1}

\KwIn{The data stream $\mathcal{S}$ with unknown length $|\mathcal{S}|=\sum_{i=1}^{N}N_{\mathcal{T}_i}^{\text{tr}}$; the $k$\textsuperscript{th} sample $(\mathbf{X}, \mathbf{Y})_k$ from $\mathcal{S}$; the long-term buffer $\mathcal{M}$ with the buffer size $|\mathcal{M}|$; the temporal buffer $\mathcal{M}_\text{tmp}$.}
\KwOut{The updated $\mathcal{M}$, where the $r$\textsuperscript{th} sample stored in $\mathcal{M}$ is denoted as $\mathcal{M} (r)$; the updated temporal buffer $\mathcal{M}_\text{tmp}$}

$\mathcal{M} \gets \emptyset$\tcp*[r]{Initialization of the long-term buffer.}
$\mathcal{M}_\text{tmp} \gets \emptyset$ \tcp*[r]{Initialization of the temporal buffer.}

\For{$k$ \textbf{in} range(1, $|\mathcal{S}|$)}{
    $\mathcal{M}_\text{tmp} \leftarrow (\mathbf{X}, \mathbf{Y})_k$ \tcp*[r]{Update the temporal buffer.}
    \eIf{$k \leq |\mathcal{M}|$}{
        $\mathcal{M}(k) \gets (\mathbf{X}, \mathbf{Y})_k$\tcp*[r]{Add sample before full.}
        $\mathcal{M} \gets \mathcal{M} \cup \mathcal{M}(k)$\tcp*[r]{Update the long-term buffer.}
    }{
        $r \sim \text{Uniform}(1, k)$\;
        \If{$r \leq |\mathcal{M}|$}{
            $\mathcal{M}(r) \gets (\mathbf{X}, \mathbf{Y})_k$\tcp*[r]{Update the long-term buffer.}
        }
    }
}
\end{algorithm}

Different from the long-term buffer that stores previously learned knowledge, the other sub-module within the memory buffer, i.e., the temporal buffer $\mathcal{M}_\text{tmp}$, saves one batch of lastly encountered training samples in the interim until the next optimization. These temporarily buffered samples from the current task $\mathcal{T}_c$ serve as reference baselines for similarity computation in the selective memory rehearsal phase.

\subsection{Selective Memory Rehearsal}
The CL objective \eqref{eq_dilemma} presents the adaptation with data in the current task while retaining learned knowledge from past tasks. Under assumptions of CL, only the training samples within the current task $\mathcal{T}_c$ and memory buffer $\mathcal{M}$ are available. In such a case, the objective \eqref{eq_dilemma} can be further formulated as a constraint optimization problem:
\begin{equation}
    \begin{aligned}
        \boldsymbol{\theta}_{1:c}^{*} &= \arg\min_{\boldsymbol{\theta}_{1:c}} \ell\left(f_{\boldsymbol{\theta}_{c}}\left( \mathbf{X}^{\mathcal{D}_{\mathcal{T}_c}}\right), \mathbf{Y}^{\mathcal{D}_{\mathcal{T}_c}}\right) \\
        &  \text{s.t.} \ \ell (f_{\boldsymbol{\theta}_{c}}, \mathcal{M}) \le \ell (f_{\boldsymbol{\theta}_{c-1}}, \mathcal{M}), \forall c\in[1,...,N]
    \end{aligned}
    \label{eq_agem_constraint}
\end{equation}
where $\ell(f_{\boldsymbol{\theta}_c}, \mathcal{M})$ denotes the loss on the memory buffer when the model is trained with task $\mathcal{T}_c$. Thus, the constraint could enhance the memory stability by limiting the loss increment on previously learned knowledge. As suggested by~\cite{lopez2017gradient-gem}, the constraint in \eqref{eq_agem_constraint} can be rephrased in a gradient format:
\begin{equation}
\left \langle \frac{\partial \ell(f_{\boldsymbol{\theta}}(\mathbf{X}^{\mathcal{D}_{\mathcal{T}_c}}) , \mathbf{Y}^{\mathcal{D}_{\mathcal{T}_c} }    )}{\partial \boldsymbol{\theta}}, \frac{\partial \ell(f_{\boldsymbol{\theta}},\mathcal{M})}{\partial \boldsymbol{\theta } }   \right \rangle \ge 0 \label{eq_gradient_constraint}
\end{equation}
where $\langle \cdot,\cdot \rangle$ is the inner product function for the gradients.

The constraint \eqref{eq_gradient_constraint} indicates that using samples from the memory buffer with loss gradient diverse from the current one might harm the adaptation to task $\mathcal{T}_c$. In other words, the learning plasticity could be impeded by re-learning memory samples that are different from training samples $(\mathbf{X}^{\mathcal{D}_{\mathcal{T}_c}}, \mathbf{Y}^{\mathcal{D}_{\mathcal{T}_c}})$. On the other hand, the strong learning plasticity means the CL model can utilize learned knowledge from past tasks to enhance its performance in the current task. Motivated by this idea, we unlock the learning plasticity of the motion forecasting model by replaying buffer samples that are similar to the currently learned ones.

\begin{figure}[tp]
      \centering
      \includegraphics[scale=1.0]{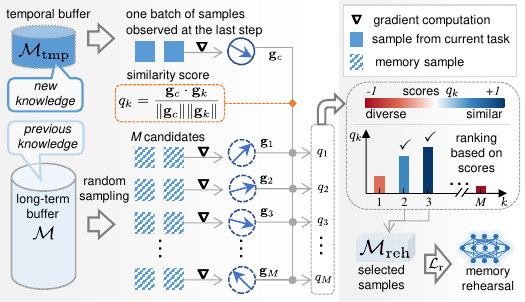}
      \captionsetup{font={small}}
      \caption{Selective memory rehearsal for learning plasticity.}
      \label{fig_selective}
\end{figure}

Specifically, gradient cosine similarity is employed to score the memory samples, measuring their degree of similarity to the current task. As illustrated in Fig.~\ref{fig_selective}, with the help of the temporal buffer $\mathcal{M}_{\text{tmp}}$, the loss gradient of the lastly observed sample from the current task $\mathcal{T}_c$ is recorded, and denoted as $\mathbf{g}_{c}$. Let $B$ denote the batch size for training. Before the next update step, $M \ge 2B$ samples from the memory buffer $\mathcal{M}$ are randomly selected as candidates for memory rehearsal, and we compute their corresponding loss gradients $\{\mathbf{g}_k\}_{k=1}^{M}$. Then, we score these $M$ samples by measuring the cosine similarity $q_{k}$ between between $\{\mathbf{g}_k\}_{k=1}^{M}$ and $\mathbf{g}_{c}$:

\begin{equation}
    q_k = \frac{\mathbf{g}_c \cdot \mathbf{g}_k}{\|\mathbf{g}_c\| \|\mathbf{g}_k\|}
\end{equation}

The range of the score $q_k$ could vary from -1 to 1, where the minimum value -1 indicates that the direction of the gradient $\mathbf{g}_k$ is opposite to the gradient of the last observed samples, $\mathbf{g}_c$. In contrast, if the two gradients are the same, the score $q_k$ equals 1. Since a larger score demonstrates higher similarity between the memory sample and the current task, we rank $M$ candidates based on the scoring results. The $B$ samples that have higher scores are used for the memory rehearsal, denoted as $\mathcal{M}_\text{reh}$. Built on these selected memory samples, the proposed selective memory rehearsal is conducted through the forward propagation by computing the following loss:
\begin{equation}
    \mathcal{L}_\text{r} = \mathbb{E}_{(\mathbf{X,Y})\sim \mathcal{M}_{\text{reh}}}(\ell(f_{\boldsymbol{\theta}}(\mathbf{X}),\mathbf{Y}))\label{eq_reh}
\end{equation}

On the one hand, the selected memory samples have similar gradients to the lastly encountered samples from the current task, where $\mathcal{M}_{\text{reh}}$ might contain related knowledge for the adaptation to $\mathcal{T}_c$. Thus, the selected memory rehearsal using $\mathcal{M}_{\text{reh}}$ has the potential to improve the performance in the current $\mathcal{T}_c$. It can be considered as a variant of a data augmentation strategy in the context of online CL settings, where the current training data is augmented with a batch of learned samples. On the other hand, unlike using new data to enhance plasticity, the selective memory rehearsal is a type of mild and friendly strategy to memory stability since the reused data originally comes from past tasks. Finally, the total loss is the combination of the loss on current samples $(\mathbf{X}^{\mathcal{D}_{\mathcal{T}_c}}, \mathbf{Y}^{\mathcal{D}_{\mathcal{T}_c}})$ and the rehearsal loss \eqref{eq_reh}:

\begin{equation}
    \mathcal{L}_{\text{total}} = \ell(f_{\boldsymbol{\theta}_c}(\mathbf{X}^{\mathcal{D}_{\mathcal{T}_c}}), \mathbf{Y}^{\mathcal{D}_{\mathcal{T}_c}}) + \mathcal{L}_\text{r}
\end{equation}

\subsection{Gradient Projection Constraint}

The selective memory rehearsal module is inspired by the gradient format goal formulated in \eqref{eq_agem_constraint} and \eqref{eq_gradient_constraint}, incorporating the rehearsal loss $\mathcal{L}_\text{r}$ into the total loss for the enhancement of learning plasticity. Synergistically, the gradient projection constraint module aims at handling the other side of the dilemma, i.e., memory stability. The core idea is to set an inequality constraint for the gradient descent-based optimization on the total loss.

Formally, let $\mathbf{g}$ denote the gradient of total loss, and let $\mathbf{g}_\mathcal{M}$ denote the gradient of average loss on the long-term buffer. We replace the current loss in \eqref{eq_agem_constraint} as the total loss, where the constraint limits the average loss increment on previously learned knowledge within the long-term buffer $\mathcal{M}$. The optimization with the gradient-format inequality constraint can be represented as:
\begin{equation}
    \begin{aligned}
        \boldsymbol{\theta}_{1:c}^{*} &= \arg\min_{\boldsymbol{\theta}_{1:c}} \mathcal{L}_{\text{total}} \\
        &  \text{s.t.} \ \langle \mathbf{g}, \mathbf{g}_{\mathcal{M}} \rangle \ge 0
    \end{aligned}
\end{equation}

\begin{figure}[tp]
      \centering
      \includegraphics[scale=1.0]{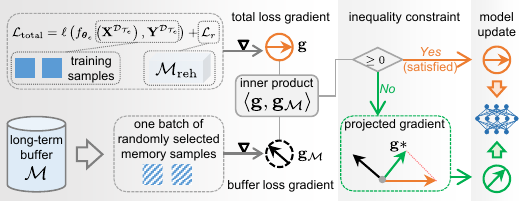}
      \captionsetup{font={small}}
      \caption{Gradient projection constraint for memory stability.}
      \label{fig_gradient}
\end{figure}

As depicted in Fig.~\ref{fig_gradient}, when the constraint is satisfied, the motion forecasting model is updated via gradient descent based on $\mathbf{g}$ directly. Otherwise, we project $\mathbf{g}$ to the closest in L2 norm gradient $\tilde{\mathbf{g}}$:
\begin{equation}
    \begin{aligned}
        \tilde{\mathbf{g}}^{*} &= \arg\min_{\tilde{\mathbf{g}}} \frac{1}{2} \left\| \mathbf{g} - \tilde{\mathbf{g}} \right\|^{2}_{2} \\
        &\quad \text{s.t.} \quad \tilde{\mathbf{g}}^{\top} \mathbf{g}_{\mathcal{M}} \ge 0
    \end{aligned}
    \label{eq_opt}
\end{equation}

The $\tilde{\mathbf{g}}^{*}$ is the solution to \eqref{eq_opt}, which satisfies the constraint, and is used for model update. In implementations, $\mathbf{g}_{\mathcal{M}}$ is computed based on a batch of samples randomly selected from the long-term buffer $\mathcal{M}$. The Lagrangian function for the optimization problem \eqref{eq_opt} can be formulated as:
\begin{equation}
    {L}(\tilde{\mathbf{g}}, \lambda) = \frac{1}{2} \| \mathbf{g} - \tilde{\mathbf{g}} \|_2^2 - \lambda (\tilde{\mathbf{g}}^\top \mathbf{g}_{\mathcal{M}})
\end{equation}
where \(\lambda \geq 0\) is the Lagrangian multiplier. To obtain the optimal solution, we first calculate the derivative of the Lagrangian function $\mathcal{L}$ with respect to $\tilde{\mathbf{g}}$:
\begin{equation}
    \nabla_{\tilde{\mathbf{g}}} L = \tilde{\mathbf{g}} -\mathbf{g} - \lambda \mathbf{g}_{\mathcal{M}}\label{eq_dv}
\end{equation}
Then, we set $\nabla_{\tilde{\mathbf{g}}} L$ in \eqref{eq_dv} to zero for the optimality condition based on the first-order optimality condition:
\begin{equation}
\tilde{\mathbf{g}} = \mathbf{g} + \lambda \mathbf{g}_{\mathcal{M}}\label{eq_tilde_g}
\end{equation}

Meanwhile, when the constraint in \eqref{eq_opt} is not satisfied, according to Karush-Kuhn-Tucker (KKT) complementary slackness, we obtain: 
\begin{equation}
    \tilde{\mathbf{g}}^\top \mathbf{g}_{\mathcal{M}} = 0 \label{eq_kkt}
\end{equation}
Substituting \eqref{eq_tilde_g} into \eqref{eq_kkt}, we have \((\mathbf{g} + \lambda \mathbf{g}_{\mathcal{M}})^\top \mathbf{g}_{\mathcal{M}} = 0\), leading to \(\lambda = -\frac{\mathbf{g}^\top \mathbf{g}_{\mathcal{M}}}{\| \mathbf{g}_{\mathcal{M}} \|_2^2}\). Therefore, the solution, i.e., the projected gradient $\tilde{\mathbf{g}}^{*}$ is obtained as:
\begin{equation}
    \tilde{\mathbf{g}}^{*} = \mathbf{g} - \frac{\mathbf{g}^\top \mathbf{g}_{\mathcal{M}}}{\| \mathbf{g}_{\mathcal{M}} \|_2^2} \mathbf{g}_{\mathcal{M}}
\end{equation}

Finally, the solution for the gradient projection constraint is represented as:
\begin{equation}
\tilde{\mathbf{g}}^* = 
\begin{cases} 
\mathbf{g}, & \text{if } \mathbf{g}^\top \mathbf{g}_{\mathcal{M}} \geq 0, \\
\mathbf{g} - \dfrac{\mathbf{g}^\top \mathbf{g}_{\mathcal{M}}}{\|\mathbf{g}_{\mathcal{M}}\|_2^2} \mathbf{g}_{\mathcal{M}}, & \text{if } \mathbf{g}^\top \mathbf{g}_{\mathcal{M}} < 0.
\end{cases}
\end{equation}
where the gradient $\tilde{\mathbf{g}}^{*}$ is used to update the DNN model parameters $\boldsymbol{\theta}$.

\section{Experiments}\label{Section-V}

This study proposes SyReM, an online CL method, for DNN-based motion forecasting models to address the stability-plasticity dilemma. Leveraging naturalistic driving datasets collected from diverse scenarios, we construct sequential tasks for DNN-based motion forecasting. The memory stability and learning plasticity of SyReM are comprehensively evaluated using various metrics. Compared with non-CL and CL baselines, the performance and effectiveness of SyReM are systematically assessed from multiple perspectives, including its ability to resolve the stability-plasticity dilemma and overall CL performance. Furthermore, beyond the dilemma, zero-shot generalization of methods to handle unseen tasks is also investigated in this section.

\begin{figure*}[tp]
      \centering
      \includegraphics[scale=1.0]{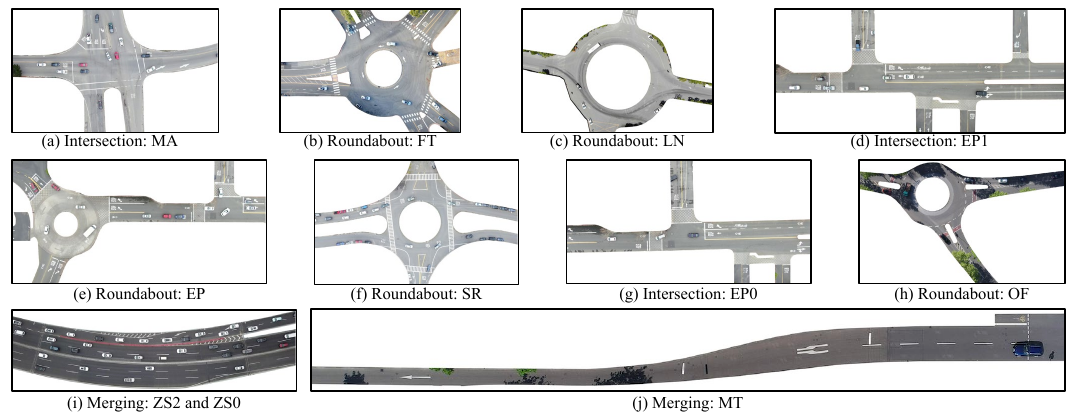}
      \captionsetup{font={small}}
      \caption{Scenarios of INTERACTION dataset~\cite{zhan2019interaction-dataset} used in the experiments.}
      \label{fig_dataset_bev}
\end{figure*}

\begin{figure}[tp]
      \centering
      \includegraphics[scale=1.0]{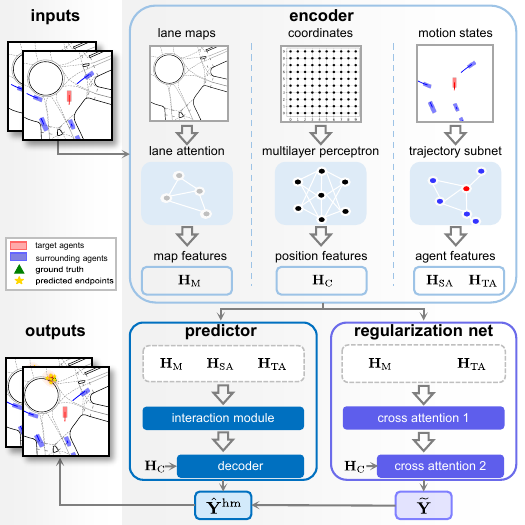}
      \captionsetup{font={small}}
      \caption{UQnet: The DNN-based motion forecasting model implemented in the experiments.}
      \label{fig_uqnet}
\end{figure}

\subsection{Experimental Settings}
\subsubsection{Dataset and Basic Model} INTERACTION is a widely used dataset for CL in motion forecasting~\cite{zhan2019interaction-dataset}. Eleven sub-datasets within INTERACTION are used in our experiments to construct sequential tasks for motion forecasting. These sub-datasets are collected from different scenarios, as shown in Table~\ref{table_datasets}. The bird-eye-view's photos of corresponding real-world scenarios are demonstrated in Fig.~\ref{fig_dataset_bev}.

The basic model implemented in the experiments is the uncertainty quantification network (UQnet)~\cite{li2024UQnet}, which represents the state-of-the-art (SOTA) for vehicle motion forecasting in the open INTERPRET challenge\footnote{INTERPRET: Interaction-dataset-based prediction challenge, single-agent track, organized by ICCV Competition. Available at: \url{https://challenge.interaction-dataset.com/leader-board/}. Last accessed on July 30\textsuperscript{th}, 2025.}. The SyReM and all compared methods are integrated into the UQnet framework. As illustrated in Fig.~\ref{fig_uqnet}, the UQnet architecture combines multiple feature encoding modules to process spatial and motion information. Specifically, an attention-based module encodes road connectivity features from the map (denoted as $\mathbf{H}_{\text{M}}$), while a multi-layer perceptron processes mesh grid coordinates to obtain position features ($\mathbf{H}_{\text{C}}$). The motion states of SAs and the TA are encoded as $\mathbf{H}_{\text{SA}}$ and $\mathbf{H}_{\text{TA}}$, respectively. These feature representations are subsequently fed into a regularization network and a DenseTNT-based predictor~\cite{gu2021densetnt}. 

By incorporating a regularization term derived from the focal loss~\cite{2020focal_loss}, the UQNet forecasts the TA's future positions as a two-dimensional spatial distribution represented by a mesh-grid heatmap $\hat{\mathbf{Y}}^{\text{hm}}$. Each grid value $\hat{\mathbf{Y}}_{i,j}^{\text{hm}}$ (where $i \in [1, h]$, $j \in [1, w]$ for an $h \times w$ heatmap) corresponds to the probability of the TA occupying that particular cell. Finally, $W$ predicted endpoints are extracted using a local-maximum sampling strategy. For comprehensive architectural details, we refer readers to~\cite{li2024UQnet}.

\begin{table}[bp]
    \centering
    \captionsetup{font={small}}
    \caption{Information Of Scenarios Used In Experiments}
    \begin{tabular}{c c c c c c}
    \toprule
 \makecell{Task \\ID} & \makecell{Scenario \\ Type} & \makecell{Sub-dataset \\Name} & \makecell{Video \\ Length (min)} & \makecell{Training\\ Cases} & \makecell{Test \\ Cases} \\ \midrule
$\mathcal{T}_1$ & Intersection & MA  & 107.37  & 33,456  & 1,178 \\
$\mathcal{T}_2$ & Roundabout & FT  & 207.62  & 66,256  & 2,452 \\
$\mathcal{T}_3$ & Roundabout & LN  & 227.00  & 3,400   & 238   \\
$\mathcal{T}_4$ & Merging & ZS2 & 94.62   & 15,400  & 1,043 \\
$\mathcal{T}_5$ & Roudabout & OF  & 55.04   & 7,904   & 568   \\
$\mathcal{T}_6$ & Intersection &EP0 & 37.30   & 9,136   & 556   \\
$\mathcal{T}_7$ & Merging & ZS0 & 94.62   & 35,512  & 1,084 \\
$\mathcal{T}_8$ & Intersection & EP & 37.30   & 12,256  & 433 \\
$\mathcal{T}_9$ & Merging & MT & 37.93   & 5,380  & 343 \\
$\mathcal{T}_{10}$ & Roundabout & SR & 40.09   & 13,748  & 552 \\
$\mathcal{T}_{11}$ & Intersection & EP1 & 37.30   & 10,048  & 297 \\
    \bottomrule
    \end{tabular}
    \label{table_datasets}
\end{table}

\subsubsection{Baselines and Implementation Details}
We compare our proposed method with the following baselines to reveal the effectiveness and advantages of SyReM. All the comparison methods are implemented on the UQnet, and the detailed configurations of the models are presented as follows:
\begin{itemize}
\item \textbf{Vanilla}: This is the original UQnet without implementing any CL methods. Serving as a non-CL baseline, Vanilla represents the basic capabilities of the DNN-based motion forecasting in the online CL tasks. Vanilla is trained on the same data stream with the same hyperparameters as the proposed method, making sure comparisons are fair.  
\item \textbf{Vanilla-GP}: Built upon the same network architecture as Vanilla, Vanilla-GP employs the proposed gradient projection constraint module with the long-term buffer to strengthen memory stability. It is equivalent to the proposed method without the temporal buffer or selective memory rehearsal module, and adopts identical training settings and hyperparameters. This baseline serves to represent CL strategies that primarily focus on maintaining memory stability while not incorporating mechanisms for enhancing learning plasticity.
\item \textbf{JoTr}: The UQnet with joint training, where all the training data are available simultaneously. Note that it violates the fundamental assumption of CL, as CL typically assumes that data arrives incrementally and historical data cannot be fully accessible during new task learning.
\item \textbf{SyReM-R}: This baseline is used in the ablation study. The core functional module of the proposed SyReM has been modified to evaluate the effectiveness of the selective memory rehearsal component. Specifically, the similarity-based memory selection strategy is replaced with a naive random sampling mechanism. All other settings, including the training tasks, network architecture, buffer size, and hyperparameters, remain identical to those used in the proposed method.
\item \textbf{SyReM (ours)}: This is the proposed online CL method. It is designed to effectively handle sequential data and mitigate catastrophic forgetting, enabling DNN-based motion forecasting models to adaptively learn new tasks over time while retaining knowledge from previous ones.
\end{itemize}

All the models are trained with a $1\times10^{-3}$ learning rate, and Adam is adopted as the optimizer~\cite{kingma2014adam}. Training and testing cases in the experiments are detailed in Table~\ref{table_datasets}. The memory buffer size $|\mathcal{M}|$ for the compared CL methods, including Vanilla-GP, SyReM-R, and SyReM, is set to 1,000 cases. This buffer size accounts for 0.47\% of the total training cases across the 11 tasks, aligning with the CL assumption on computational resources specified in \eqref{eq_data_limit}. Besides, the training batch size $B$ is set to 8. To verify that the proposed SyReM can achieve strong performance with minimal data cost, we enforce a minimum number of candidates $M\ge 2B$ in the selective memory rehearsal. Accordingly, $M$ is set to $2B$, which equals 16. In hyperparameters of motion forecasting, the observation time $t_\text{obs}$ is 1 s, and the prediction horizon $t_\text{pred}$ is 3 s. Following the study~\cite{li2024UQnet}, the number of sampled predicted endpoints $W$ is 6. The experiments are executed on a Linux-based server equipped with an AMD EPYC-7763 CPU and an NVIDIA GeForce RTX 4090 GPU.

\subsection{Evaluation Metrics}

\begin{figure}[tp]
      \centering
      \includegraphics[scale=1.0]{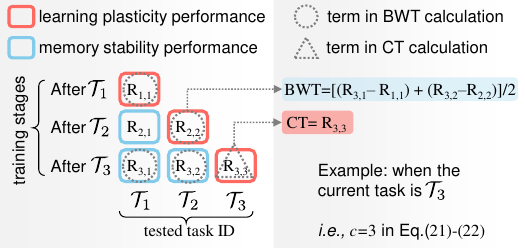}
      \captionsetup{font={small}}
      \caption{A schematic of metrics calculation. Note that $\text{FDE}_{i,j}$ and $\text{MR}_{i,j}$ are denoted as $\text{R}_{i,j}$ uniformly. In this example, the total number $N$ is 3, and the current task $\mathcal{T}_c$ incrementally increases from $\mathcal{T}_1$ to $\mathcal{T}_3$.}
      \label{fig_metrics}
\end{figure}

Minimum final displacement error (FDE) and miss rate (MR) are widely used metrics in motion forecasting~\cite{li2024UQnet, gu2021densetnt, 2024-goal-traj-nnls}. As the basic metrics adopted in this study, the FDE measures the minimum Euclidean distance between the $W$ predicted endpoints and the ground truth ${\bf{Y}}$ over each sample:
\begin{equation}
    \text{FDE} =\min_{k\in\left \{ 1,...,W \right \} }  {\left\| {{\bf{\hat {\bf{Y}}}}_{k} - {\bf{Y}}} \right\|}_2 \label{eq_fde_sample}
\end{equation}

MR measures the percentage of predicted endpoints that are out of a given lateral or longitudinal area of the ground truth. As implemented in \cite{li2024UQnet}, the lateral threshold of the given area is 1 $\rm{m}$, and the longitudinal threshold is determined by the piece-wise function depending on the velocity $v$ of the TA:

\begin{equation}
th_{\text{MR}}(v) = \left\{ \begin{array}{l}
1, v < 1.4 \\
1 + \frac{{v - 1.4}}{{11 - 1.4}}, 1.4 \le v \le 11\\
2, v > 11
\end{array} \right.
\end{equation}
where the unit of the longitudinal threshold is $\rm{m}$, and the unit of the velocity $v$ is $\rm{ms^{-1}}$. Let $W^{\text{out}}$ denote the number of predicted endpoints that are out of the given area within a case. The MR of one test case is calculated as:
\begin{equation}
    \text{MR}= \frac{W^{\text{out}}} W \times 100\% \label{eq_mr_sample}
\end{equation}
The FDE and MR within one task are average results of \eqref{eq_fde_sample} and \eqref{eq_mr_sample} over the corresponding test set. Since FDE and MR both measure the error, the smaller values of these two metrics represent better performance. 

As described in Section~\ref{Section-III}, sequential tasks are constructed to validate the CL performance, where the motion forecasting model is tested in all tasks after learning a new task. Let $\text{FDE}_{i,j}$ and $\text{MR}_{i,j}$ denote the FDE and MR within the test set of task $\mathcal{T}_{j}$ after the model learns the task $\mathcal{T}_i$. The metrics for quantifying the memory stability and learning plasticity are built based on $\text{FDE}_{i,j}$ and $\text{MR}_{i,j}$ in the following.
\subsubsection{Measurement of Memory Stability}
The FDE-based backward transfer (BWT) and MR-based BWT are used to quantify the memory stability, measuring the influence that learning the current task $\mathcal{T}_c$ has on the performance of previously learned tasks $\mathcal{T}_i, \forall i<c$:
\begin{subequations}
    \label{eqs-bwt}
    \begin{align}
        \text{FDE-BWT} = \frac{1}{c-1} \sum_{i=1}^{c-1}\left (\text{FDE}_{c,i}-\text{FDE}_{i,i}  \right ) \\
        \text{MR-BWT} = \frac{1}{c-1} \sum_{i=1}^{c-1}\left (\text{MR}_{c,i}-\text{MR}_{i,i}  \right )
    \end{align}
\end{subequations}

Note that since the BWT is calculated based on FDE or MR, the smaller FDE-BWT or MR-BWT represents the better memory stability, indicating less catastrophic forgetting.

\subsubsection{Measurement of Learning Plasticity}
The ability to acquire new knowledge by training on newly available data reveals the learning plasticity~\cite{lyle2023understanding}. Thus, the performance in the current task $\mathcal{T}_c$, denoted as FDE-CT and MR-CT, demonstrates the learning plasticity:
\begin{subequations}
    \label{eqs-ct}
    \begin{align}
        \text{FDE-CT}= \text{FDE}_{c,c} \\
        \text{MR-CT} = \text{MR}_{c,c}
    \end{align}
\end{subequations}
where FDE-CT and MR-CT evaluate the model performance in the test set of task $\mathcal{T}_c$ after it is sequentially trained by samples from $\mathcal{T}_1$ to $\mathcal{T}_c$.

As depicted in Figure~\ref{fig_metrics}, an example presents the calculation of the abovementioned metrics. For convenience, the $\text{FDE}_{i,j}$ and $\text{MR}_{i,j}$ are denoted as $\text{R}_{i,j}$ uniformly. The horizontal axis refers to the evaluation task ID, and the vertical axis demonstrates the training stages. The FDE-BWT and MR-BWT in \eqref{eqs-bwt} are simply denoted as BWT. Similarly, FDE-CT, MR-CT in \eqref{eqs-ct} are denoted as CT. 

\subsubsection{Overall Performance Measurement}
The overall performance is evaluated by testing the model in a joint test set $\Omega_{1:N}$, which combines samples from test sets of all tasks $\{\mathcal{T}_{i}\}_{i=1}^N$~\cite{kim2023dilemma-CVPR}. The model is incrementally tested in the same joint test set $\Omega_{1:N}$ after learn a task. Since $\Omega_{1:N}$ contains all testing samples from learned and unseen tasks, the averaged values of FDE and MR over the joint test set $\Omega_{1:N}$ are denoted as $\text{FDE-JT}$ and $\text{MR-JT}$, respectively.

\subsection{Experimental Results and Analysis}\label{Section-V-C}

\begin{figure}[tp]
    \centering
   
        \includegraphics[scale=1.0]{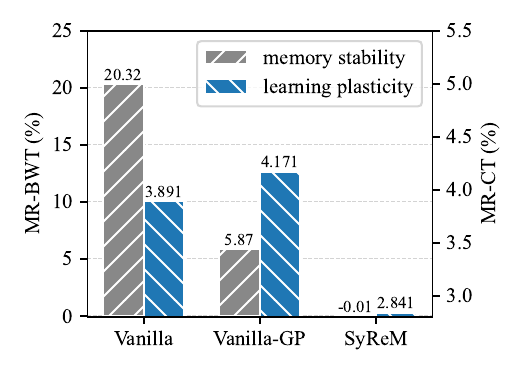}
    \captionsetup{font={small}}
    \caption{Averaged performance between memory stability and plasticity. The memory stability is measured by MR-BWT, and the learning plasticity is measured by MR-CT.}
    \label{fig_SP_bars_combined}
\end{figure}

\subsubsection{Escape from Stability-Plasticity Dilemma}

The motion forecasting model is incrementally trained by the one-pass online data stream for settings including Vanilla, Vanilla-GP, and SyReM. The model is evaluated in all test sets of $\{\mathcal{T}_{i}\}_{i=1}^N$ each time it finishes observing the training samples of one task $\mathcal{T}_i, \forall i\in[1, N]$. To demonstrate the stability-plasticity dilemma, the MR-BWT and MR-CT averaged in eleven experimental groups of CL tasks are compared as shown in Fig.~\ref{fig_SP_bars_combined}.

\begin{figure*}[tp]
      \centering
      \includegraphics[scale=1.0]{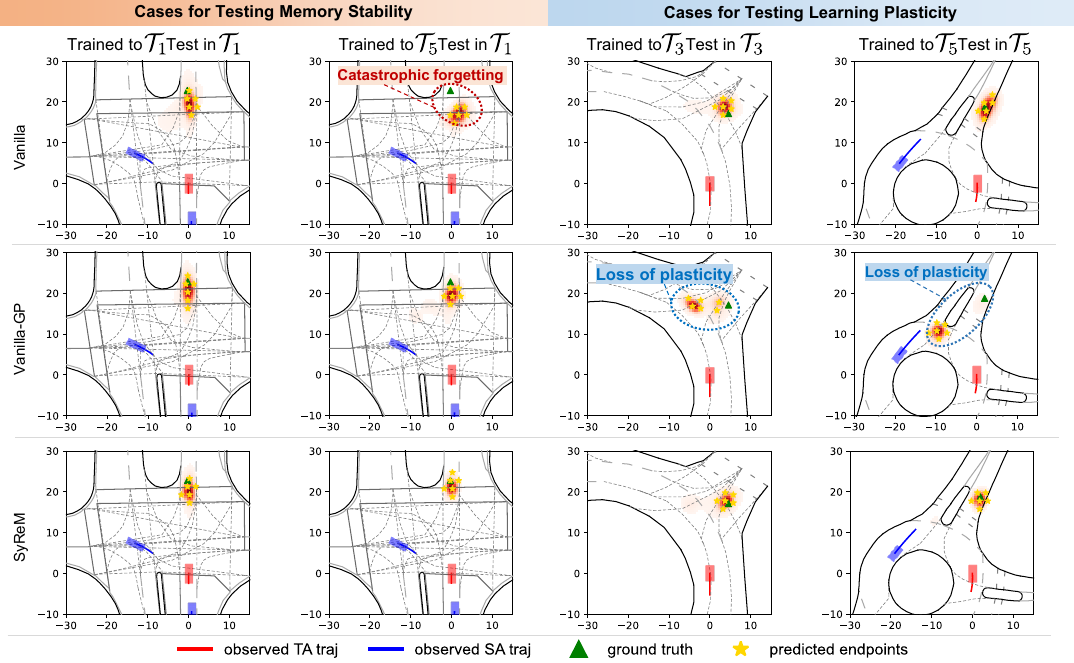}
      \captionsetup{font={small}}
      \caption{Experimental cases showing the stability-plasticity dilemma of baseline models, which is addressed by the proposed method, SyReM.}
      \label{fig_case}
\end{figure*}

MR-BWT, representing the memory stability, is colored in gray, while MR-CT is colored in blue to quantify the learning plasticity. Comparing Vanilla and Vanilla-GP, it can be found that Vanilla has significantly higher MR-BWT than Vanilla-GP. Specifically, the averaged MR-BWT of Vanilla is 20.32\% and 5.87\% for Vanilla-GP, where the MR-BWT of Vanilla is higher than Vanilla-GP by 14.45\%. These experiments show that Vanilla suffers a severe catastrophic forgetting, which was alleviated by applying the gradient projection constraint in Vanilla-GP. Despite Vanilla-GP improving the memory stability of Vanilla, the learning plasticity of the model is impeded, where Vanilla has a lower average MR-CT than Vanilla-GP. By contrast, the proposed SyReM improves both the memory stability and learning plasticity of Vanilla simultaneously. For memory stability, SyReM achieves the lowest MR-BWT as -0.01\% among the compared methods. Such a negative MR-BWT indicates that learning the new data with SyReM does not harm the memory stability, but helps improve the motion forecasting performance in past tasks. Meanwhile, SyReM reduces the MR-CT of Vanilla by 27\%, significantly improving the learning plasticity of this non-CL baseline. Detailed experimental results for memory stability (measured by FDE-BWT and MR-BWT) and learning plasticity (measured by FDE-CT and MR-CT) in each training stage are reported in Table~\ref{table_bwt} and Table~\ref{table_ct}, respectively. We find that the SyReM has the lowest FDE-BWT and MR-BWT in all training stages among the compared methods and has the lowest MR-CT in most groups. These experimental results demonstrate that the compared CL method got trapped in the stability-plasticity dilemma. Although the catastrophic forgetting of the non-CL model (i.e., Vanilla) can be alleviated by Vanilla-GP, the learning plasticity could be harmed by its stability-targeted CL strategy. The proposed SyReM escapes the dilemma by enhancing the performance on both previous tasks and the current tasks.

\begin{table}[htbp]
\centering
\small
\caption{Memory Stability Performance Measured by FDE-BWT (m) / MR-BWT (\%)}
\label{table_bwt}
\begin{tabular*}{\columnwidth}{@{\extracolsep{\fill}} c *{2}{c} c }
\toprule
\textbf{Tasks}   & \textbf{Vanilla} & \textbf{Vanilla-GP} & \textbf{SyReM} \\
\midrule
$\mathcal{T}_1 -\mathcal{T}_2$ & 0.16 / 3.90 & 0.24 / 3.48 & \textbf{-0.11} / \textbf{-4.67} \\
$\mathcal{T}_1 -\mathcal{T}_3$ & 0.41 / 9.55 & 0.75 / 19.01 & \textbf{-0.09} / \textbf{-3.03} \\
$\mathcal{T}_1 -\mathcal{T}_4$ & 1.80 / 45.91 & 0.02 / 1.32 & \textbf{0.00} / \textbf{1.17} \\
$\mathcal{T}_1 -\mathcal{T}_5$ & 1.92 / 30.77 & 0.26 / 6.07 & \textbf{-0.04} / \textbf{-1.66} \\
$\mathcal{T}_1 -\mathcal{T}_6$ & 1.68 / 25.84 & 0.08 / 1.42 & \textbf{-0.01} / \textbf{-0.77} \\
$\mathcal{T}_1 -\mathcal{T}_7$ & 2.18 / 38.24 & 0.31 / 8.06 & \textbf{0.12} / \textbf{3.16} \\
$\mathcal{T}_1 -\mathcal{T}_8$ & 1.77 / 20.47 & 0.30 / 7.22 & \textbf{0.07} / \textbf{1.76} \\
$\mathcal{T}_1 -\mathcal{T}_9$ & 0.55 / 11.07 & 0.23 / 4.56 & \textbf{0.09} / \textbf{1.89} \\
$\mathcal{T}_1 -\mathcal{T}_{10}$ & 0.68 / 7.34 & 0.17 / 3.77 & \textbf{0.06} / \textbf{0.89} \\
$\mathcal{T}_1 -\mathcal{T}_{11}$ & 0.83 / 10.10 & 0.18 / 3.80 & \textbf{0.09} / \textbf{1.18} \\
\bottomrule
\end{tabular*}
\end{table}

\begin{table}[htbp]
\centering
\small
\caption{Learning Plasticity Performance measured by FDE-CT (m) / MR-CT (\%)}
\label{table_ct}
\begin{tabular*}{\columnwidth}{@{\extracolsep{\fill}} c *{3}{c} }
\toprule
\textbf{Tasks}  & \textbf{Vanilla} & \textbf{Vanilla-GP} & \textbf{SyReM} \\
\midrule
$\mathcal{T}_1 -\mathcal{T}_2$   & \textbf{0.90} / \textbf{4.85}  & 1.05 / 8.73  & 0.97 / 5.14 \\
$\mathcal{T}_1 -\mathcal{T}_3$  & \textbf{0.79} / 1.68  & 1.08 / 8.82  & 0.80 / \textbf{1.26} \\
$\mathcal{T}_1 -\mathcal{T}_4$  & 0.72 / \textbf{0.67}  & \textbf{0.65} / 0.77  & 0.71 / 0.77 \\
$\mathcal{T}_1 -\mathcal{T}_5$   & 1.14 / 10.56  & 0.98 / 5.63  & \textbf{0.96} / \textbf{3.70} \\
$\mathcal{T}_1 -\mathcal{T}_6$   & 0.76 / \textbf{0.72}  & \textbf{0.70} / 1.44  & 0.73 / \textbf{0.72} \\
$\mathcal{T}_1 -\mathcal{T}_7$   & \textbf{0.60} / 0.18  & 0.63 / \textbf{0.00}  & 0.62 / \textbf{0.00} \\
$\mathcal{T}_1 -\mathcal{T}_8$  & 0.78 / \textbf{1.16}  & 0.75 / 2.55  & \textbf{0.68} / \textbf{1.16} \\
$\mathcal{T}_1 -\mathcal{T}_9$  & 0.68 / \textbf{0.87}  & \textbf{0.67} / \textbf{0.87}  & 0.72 / 1.17 \\
$\mathcal{T}_1 -\mathcal{T}_{10}$  & 0.54 / 0.54  & 0.57 / \textbf{0.36}  & \textbf{0.50} / 0.54 \\ 
$\mathcal{T}_1 -\mathcal{T}_{11}$  & 0.71 / 2.03  & \textbf{0.67} / 2.36  & 0.68 / \textbf{1.01} \\
\bottomrule
\end{tabular*}
\end{table}

For a more intuitive demonstration, experimental cases are shown in Fig.~\ref{fig_case}. Models are sequentially trained from task $\mathcal{T}_1$ to $\mathcal{T}_5$. Within each sub-figure in Fig~\ref{fig_case}, the red and blue rectangles represent the TA and SAs, respectively. The positions of TA after 3 s (ground truth) are denoted as green triangles, and the predictions are represented as gold stars with red heatmaps. The test cases on task $\mathcal{T}_1$ at different training stages, including after learning $\mathcal{T}_1$ and $\mathcal{T}_5$, demonstrate the performance of memory stability. As illustrated in the first column from the left side of Fig.~\ref{fig_case}, all three models have accurate predictions for the $\mathcal{T}_1$ test case, after being trained in task $\mathcal{T}_1$. However, in the same test case, the distance between the predictions and the ground truth is more than 3 m for Vanilla after it was trained continually to $\mathcal{T}_5$. Such an increment of prediction errors after learning new tasks indicates that Vanilla suffers from catastrophic forgetting. We can also observe that Vanilla-GP and SyReM have relatively accurate predictions, as tested in $\mathcal{T}_1$ after being trained on $\mathcal{T}_5$, showing better memory stability than Vanilla. 

In cases for testing learning plasticity, Fig.~\ref{fig_case} illustrates test cases in $\mathcal{T}_3$ and $\mathcal{T}_5$, when the model is trained to $\mathcal{T}_3$ and $\mathcal{T}_5$, respectively. Notably, the predictions from Vanilla-GP deviate far from the ground truth in tasks $\mathcal{T}_3$ and $\mathcal{T}_5$, indicating a severe loss of learning plasticity. Compared to Vanilla-GP, both Vanilla and the proposed SyReM obtain accurate predictions in these cases. In summary, the non-CL baseline, Vanilla, suffers from catastrophic forgetting. Despite Vanilla-GP mitigating such forgetting due to its mechanism to enhance memory stability, the learning plasticity is impeded, leading to poor performance in newly encountered tasks. Comparably, SyReM escapes from such a stability-plasticity dilemma, achieving consistently accurate motion forecasting in online CL tasks.

\subsubsection{Overall Performance Evaluation}

\begin{figure}[tp]
      \centering
      \includegraphics[scale=1.0]{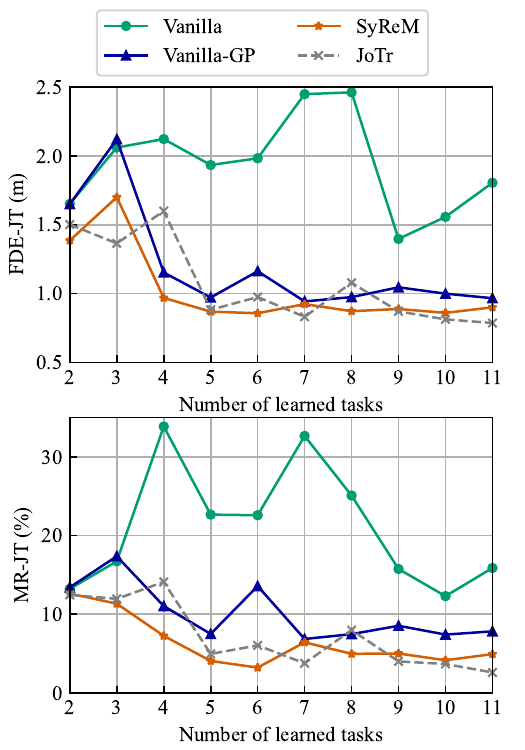}
      \captionsetup{font={small}}
      \caption{FDE-JT and MR-JT in the joint test set $\Omega_{1:11}$, which combines test data from task $\mathcal{T}_1$ to $\mathcal{T}_{11}$. The x-axis refers to the number of incrementally learned tasks.}
      \label{fig_joint_test}
\end{figure}

The overall performance of models is evaluated based on the joint test set. As demonstrated in Fig.~\ref{fig_joint_test}, FDE-JT and MR-JT of different methods are compared. The x-axis refers to the number of sequentially learned tasks, and all the values of FDE-JT and MR-JT are tested in the same test set $\Omega_{1:11}$, which combines all test sets of CL tasks $\{\mathcal{T}_{i}\}_{i=1}^{11}$. Note that the gray dash lines with cross markers represent the results of JoTr, which is trained using all of the available datasets at once. For example, instead of using the online data stream, the JoTr is trained with the dataset combined by $N' \in [2, ..., 11]$ training sets from $\{\mathcal{T}_{i}\}_{i=1}^{N'}$ when the number of learned tasks is $N'$.

From the perspective of the overall trend, the FDE-JT and MR-JT of JoTr gradually become smaller with learned tasks increases. This is reasonable since more training data is observed. Compared to JoTr, Vanilla, Vanilla-GP, and SyReM follow the training paradigm of online CL, sequentially observing the one-pass data stream. In such a challenging setting, the fluctuation of FDE-JT and MR-JT of Vanilla (green lines with dot markers) is relatively large. The FDE-JT and MR-JT are not reduced when more tasks are learned, indicating the inability of Vanilla to accumulate knowledge in online CL. Represented by the blue lines with triangle markers, Vanilla-GP has lower FDE-JT and MR-JT than Vanilla in most cases, showing the effectiveness of its CL strategy. However, the Vanilla-GP only outperformed the JoTr when the learned tasks were 4 and 8. Compared to Vaniall and Vanilla-GP, the SyReM (orange lines with star makers) has the lowest FDE-JT and MR-JT in most cases. Besides, it achieved close to or better performance than the JoTr, demonstrating its online CL capability to accumulate learned knowledge in dealing with past or unseen tasks.

\begin{figure*}[tp]
      \centering
      \includegraphics[scale=1.0]{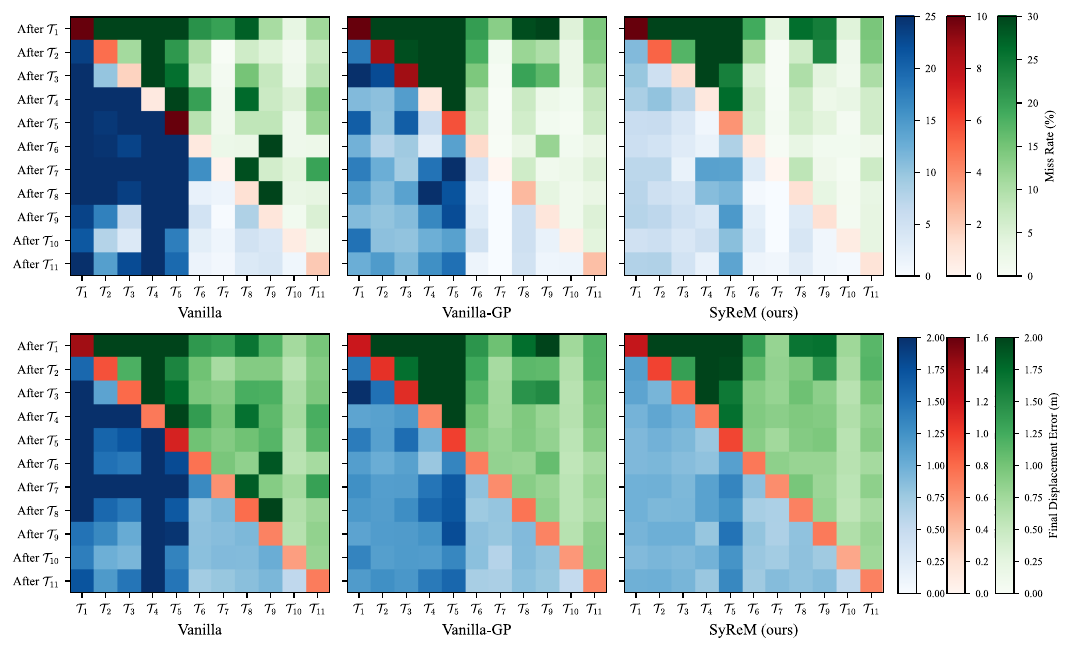}
      \captionsetup{font={small}}
      \caption{Detailed FDE (m) and MR (\%) at each training stages.}
      \label{fig_arrays}
\end{figure*}

\subsubsection{Generalization Beyond the Dilemma}
Leveraging learned knowledge to enhance performance in totally unseen tasks is also important in the online CL setting when the data is unavailable in new tasks~\cite{TPAMI2024-comprehensive-cl}. For a comprehensive evaluation, forward transfer (FWT) is used to measure the zero-shot generalization. $\forall c \le N-1$:
\begin{subequations}
    \label{eqs-fwt}
    \begin{align}
        \text{FDE-FWT} = \frac{1}{N-c}\sum_{k=c+1}^{N} \text{FDE}_{c,k} \\
        \text{MR-FWT} = \frac{1}{N-c}\sum_{k=c+1}^{N} \text{MR}_{c,k}
    \end{align}   
\end{subequations}
where $N$ is the total number of CL tasks, and FDE-FWT and MR-FWT evaluates the average performance on unseen tasks $\{\mathcal{T}_{k}\}_{k=c+1}^{N}$ after sequentially learned tasks $\{\mathcal{T}_{i}\}_{i=1}^{c}$. The smaller FDE-FWT and MR-FWT indicate better zero-shot generalization.

\begin{table}[tp]
\centering
\small
\caption{Generalization to Unseen Tasks: FDE-FWT (m) / MR-FWT (\%)}
\label{table_fwt}
\begin{tabular*}{\columnwidth}{@{\extracolsep{\fill}} c *{3}{c} }
\toprule
\textbf{Learned tasks}  & \textbf{Vanilla} & \textbf{Vanilla-GP} & \textbf{SyReM} \\
\midrule
$\mathcal{T}_1 -\mathcal{T}_2$ & 1.57 / 12.00  & 1.90 / 18.53  & \textbf{1.46} / 16.10 \\
$\mathcal{T}_1-\mathcal{T}_3$  & 1.99 / 14.32  & 2.36 / 19.31  & \textbf{1.74} / \textbf{12.67} \\
$\mathcal{T}_1 -\mathcal{T}_4$  & 1.61 / 20.53  & 1.05 / 8.81  & \textbf{0.98} / \textbf{7.37} \\
$\mathcal{T}_1 -\mathcal{T}_5$  & 0.98 / 6.73  & \textbf{0.82} / 3.71  & 0.83 / \textbf{3.99} \\
$\mathcal{T}_1 -\mathcal{T}_6$  & 1.00 / 8.72  & 0.79 / 3.70  & \textbf{0.76} / \textbf{1.43} \\
$\mathcal{T}_1 -\mathcal{T}_7$ & 1.20 / 14.66  & \textbf{0.76} / 3.67  & 0.80 / \textbf{4.35} \\
$\mathcal{T}_1 -\mathcal{T}_8$  & 1.22 / 13.85  & 0.71 / 2.55  & \textbf{0.69} / \textbf{2.38} \\
$\mathcal{T}_1 -\mathcal{T}_9$  & \textbf{0.72} / 3.08  & \textbf{0.72} / 3.09  & \textbf{0.72} / \textbf{2.05} \\
$\mathcal{T}_1 -\mathcal{T}_{10}$ & 0.80 / 1.69  & 0.86 / 4.05  & \textbf{0.76} / \textbf{3.38} \\
\bottomrule
\end{tabular*}
\end{table}

Beyond the stability-plasticity dilemma, zero-shot generalization is evaluated by testing the model in unseen tasks $\{\mathcal{T}_k\}_{k=c+1}^{N}$ after it learns task $\mathcal{T}_c$. Experimental results of FDE-FWT and MR-FWT in CL tasks are shown in Table~\ref{table_fwt}. The proposed SyReM has smaller values of FDE-FWT and MR-FWT than Vanilla and Vanilla-GP in most cases. These experimental results show that SyReM has better generalization in unseen tasks than compared baselines. Furthermore, Fig.~\ref{fig_arrays} illustrates the detailed FDE and MR values of the compared methods among previously learned tasks (colored in blue tones), current tasks (colored in red tones), and unseen tasks (colored in green tones) at each training stage. In each color tone, lighter shades represent smaller FDE or MR values, indicating better motion forecasting performance. Results colored in blue tones denote memory stability, while red tones represent learning plasticity, and green tones correspond to zero-shot generalization. Compared with Vanilla-GP and Vanilla, SyReM exhibits relatively lighter-colored regions in both red and blue tones. This confirms the aforementioned experimental results that SyReM is conducive to addressing the stability-plasticity dilemma. Besides, it can be found that after learning $\mathcal{T}_5$, both Vanilla-GP and SyReM significantly outperform Vanilla on unseen tasks. This indicates that the comprehensive retention of knowledge from historical tasks is beneficial for improving performance on test cases with unknown distributions in online CL tasks.

\subsubsection{Ablation Study}

\begin{figure}[tp]
      \centering
      \includegraphics[scale=1.0]{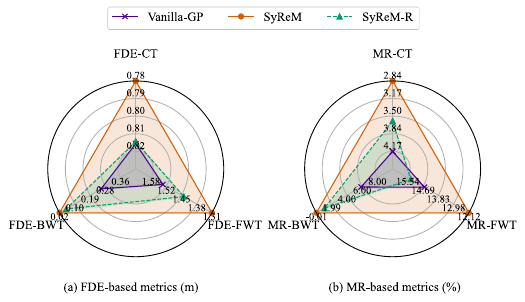}
      \captionsetup{font={small}}
      \caption{Comprehensive comparison of averaged performance measured by FDE-based metrics (a) and MR-based metrics (b).}
      \label{fig_radar}
\end{figure}

\begin{figure*}[tp]
      \centering
      \includegraphics[scale=1]{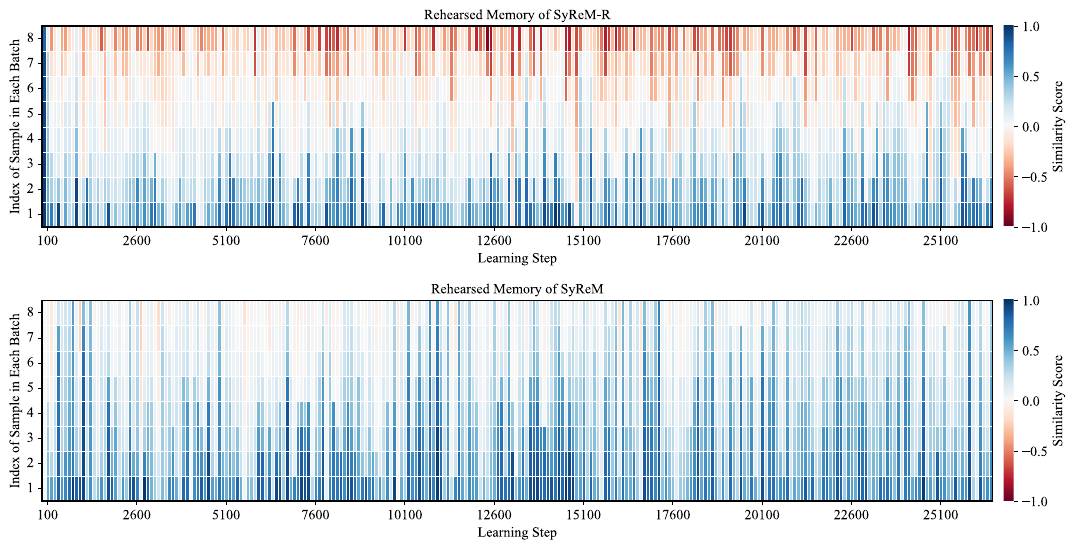}
      \captionsetup{font={small}}
      \caption{Similarity scores are used to quantify the similarity between rehearsed memory samples and current task samples observed in the last learning step, where larger scores indicate higher similarity. SyReM-R selects samples for memory rehearsal by random sampling (top). SyReM aims to maximize the similarity between rehearsed memory samples and current task knowledge (bottom).}
      \label{fig_sim}
\end{figure*}

To demonstrate the effectiveness of the proposed SyReM, the ablation model SyReM-R is compared. As described in Section~\ref{Section-IV}, a gradient-based memory strategy is developed in the SyReM. We replace this strategy with random sampling to build SyReM-R. The averaged performance is compared comprehensively, as shown in Fig.~\ref{fig_radar}. The orange lines with dot markers represent the average performance of SyReM, and green dashed lines with triangle markers represent the performance of SyReM-R. The performance of Vanilla-GP is represented by the purple line with cross markers. In Fig.~\ref{fig_radar}(a) and Fig.~\ref{fig_radar}(b), values of different metrics along each axis decrease radially outward from the center. All the metrics are averaged in 11 CL tasks.

We can find that SyReM-R achieved similar FDE-BWT and MR-BWT to SyReM, and both SyReM and SyReM-R outperform Vanilla-GP with smaller BWT-based metrics. These experimental results show that the ablation model SyReM-R has a better memory stability than Vanilla-GP, and its memory stability is close to SyReM. It is reasonable since the mechanisms of avoiding catastrophic forgetting are the same in SyReM and SyReM-R. The only difference between SyReM and SyReM-R lies in the sampling strategy from the long-term buffer in the selective memory rehearsal. The SyReM intends to replay samples that are similar to the last observed ones in the current task, while SyReM-R simply employs random sampling. Such a difference brings an obvious performance gap in learning plasticity, in which the FDE-CT and MR-CT of SyReM-R are larger than those of SyReM. Specifically, the MR-CT of SyReM is smaller than SyReM-R by 26\%. 

Diving into this core mechanism developed for enhancing learning plasticity, the similarity between the replayed samples and the lastly encountered samples in the current task is recorded for these two strategies, as shown in Fig.~\ref{fig_sim}. The similarity is measured by the cosine similarity of loss gradients in training. The x-axis refers to the training steps, and the y-axis refers to the indices of samples within a batch. The larger cosine similarity means that the replayed samples are more similar to the ones observed in the current task. As shown in Fig.~\ref{fig_sim} (bottom), most samples selected by SyReM have positive values (colored in blue tones) of cosine similarity. Nearly half replayed samples randomly selected by SyReM-R have negative values, as colored in red tones in Fig.~\ref{fig_sim} (top). Compared these recorded information with the comparison of FDE-CT and MR-CT in Fig.~\ref{fig_radar}, it can be inferred that using samples that are similar to ones in the current task for memory replay contributed to improving the learning plasticity of SyReM. These experimental results validate the effectiveness of the proposed selective memory rehearsal in SyReM.

\subsection{Discussion}

This study aims to help DNN-based motion forecasting models handle the stability-plasticity dilemma in online CL. Validated based on comprehensive metrics, experimental results show that SyReM can simultaneously enhance the memory plasticity and the learning plasticity of the DNN-based motion forecasting model. This achievement could be thanks to the developed synergy mechanisms within SyReM. By contrast, the compared methods got trapped in the stability-plasticity dilemma. Due to the lack of CL mechanisms, the non-CL baseline, i.e., Vanilla, suffers catastrophic forgetting, with the largest FDE-BWT and MR-BWT among the compared methods. Such weak memory stability impedes its applicability in the online CL paradigm. By contrast, the compared CL method, Vanilla-GP, mitigates the catastrophic forgetting of the model successfully, showing better stability than Vanilla. However, the learning plasticity is ignored in Vanilla-GP, leading to higher FDE-CT and MR-CT than Vanilla. Based on the analysis of the ablation study, we find that the selective memory rehearsal benefits the learning plasticity and also contributes to improving the memory stability. These experimental results have validated the motivation and effectiveness of the proposed method. 

Beyond the dilemma, we further provide the generalization performance of DNN-based motion forecasting toward the online CL paradigm. The CL methods, including Vanilla-GP and SyReM, both have a better zero-shot generalization than Vanilla. These results indicate that retrieving learned knowledge could potentially improve the model's robustness in unseen distributions. The ablation study further dives into the mechanisms of CL methods. Comparing the zero-shot generalization among the three methods measured by FDE-FWT (Fig.~\ref{fig_radar}(a)) and by MR-FWT (Fig.~\ref{fig_radar}(b)), SyReM has the best performance in unseen tasks. SyReM-R has a smaller FDE-FWT than Vanilla-GP, while it has a slightly larger MR-FWT than Vanilla-GP. These results indicate that random memory rehearsal might not always benefit the performance on unseen tasks in online CL. Exploring clearer underlying mechanisms for zero-shot generalization in online CL could be an encouraging future direction.

\section{Conclusion}\label{Section-VI}

This study investigates the stability-plasticity dilemma in online CL for motion forecasting. The proposed SyReM escapes the dilemma by simultaneously mitigating catastrophic forgetting and achieving accurate motion forecasting on new tasks. In the synergetic mechanisms of SyReM, the gradient projection-based inequality constraint ensures memory stability, and the selective memory rehearsal strategy enhances learning plasticity. These findings demonstrate that DNN-based models can leverage prior knowledge to adapt continually to evolving online data distributions, laying a foundation for deploying such models in dynamic real-world scenarios where training samples arrive incrementally. Beyond the dilemma, experimental results further show that SyReM outperforms baselines in zero-shot generalization across unseen tasks, with its modest yet superior improvements highlighting the potential of integrating memory rehearsal with generalization-oriented designs. Future work will explore few-shot or zero-shot learning for DNN-based intelligent systems in unseen domains while preserving incremental knowledge accumulation, aiming to strengthen generalization without sacrificing the stability-plasticity balance.

\small
\bibliographystyle{IEEEtran} 
\bibliography{references}{}

\end{document}